\documentclass[letterpaper, conference]{IEEEtran}  

\usepackage{graphicx} 
\usepackage{amsmath} 
\usepackage{amssymb}  
\usepackage{lettrine}
\usepackage{adjustbox}
\usepackage{multirow}

\usepackage[bookmarks=false]{hyperref}
\hypersetup{
colorlinks,
citecolor=black,
filecolor=black,
linkcolor=black,
urlcolor=black,
}
\usepackage{fancyhdr}

\title{Learning to Make Predictions on Graphs with Autoencoders}

\author{\IEEEauthorblockN{Phi Vu Tran}
    \IEEEauthorblockA{Strategic Innovation Group \\
    Booz $\vert$ Allen $\vert$ Hamilton \\
    San Diego, CA USA \\
    \texttt{\url{tran\_phi@bah.com}}
    }
}

\begin{document}

\maketitle
\thispagestyle{fancy}
\pagestyle{plain}

\begin{abstract}

We examine two fundamental tasks associated with graph representation learning: link prediction and semi-supervised node classification. We present a novel autoencoder architecture capable of learning a joint representation of both local graph structure and available node features for the multi-task learning of link prediction and node classification. Our autoencoder architecture is efficiently trained end-to-end in a single learning stage to simultaneously perform link prediction and node classification, whereas previous related methods require multiple training steps that are difficult to optimize. We provide a comprehensive empirical evaluation of our models on nine benchmark graph-structured datasets and demonstrate significant improvement over related methods for graph representation learning. Reference code and data are available at \url{https://github.com/vuptran/graph-representation-learning}.

\end{abstract}


\section{Introduction}

\lettrine{A}~\textsc{s} the world is becoming increasingly interconnected, graph-structured data are also growing in ubiquity. In this work, we examine the task of learning to make predictions on graphs for a broad range of real-world applications. Specifically, we study two canonical subtasks associated with graph-structured datasets: link prediction and semi-supervised node classification (LPNC). A graph is a partially observed set of edges and nodes (or vertices), and the learning task is to predict the labels for edges and nodes. In real-world applications, the input graph is a network with nodes representing unique entities, and edges representing relationships (or links) between entities. Further, the labels of nodes and edges in a graph are often correlated, exhibiting complex relational structures that violate the general assumption of independent and identical distribution fundamental in traditional machine learning \cite{Hassan:2010}. Therefore, models capable of exploiting topological structures of graphs have been shown to achieve superior predictive performances on many LPNC tasks \cite{Rossi:2012}.

We present a novel densely connected autoencoder architecture capable of learning a shared representation of latent node embeddings from both local graph topology and available explicit node features for LPNC. The resulting autoencoder models are useful for many applications across multiple domains, including analysis of metabolic networks for drug-target interaction \cite{fakhraei:2014}, bibliographic networks \cite{Sen:2008}, social networks such as Facebook (``People You May Know''), terrorist networks \cite{Zhao:2006}, communication networks \cite{Huang:2009}, cybersecurity \cite{fakhraei:2015}, recommender systems \cite{Koren:2009}, and knowledge bases such as DBpedia and Wikidata \cite{Yang:2015}.

There are a number of technical challenges associated with learning to make meaningful predictions on complex graphs:

\begin{itemize}
\item Extreme class imbalance: in link prediction, the number of known present (positive) edges is often significantly less than the number of known absent (negative) edges, making it difficult to reliably learn from rare examples;
\item Learn from complex graph structures: edges may be directed or undirected, weighted or unweighted, highly sparse in occurrence, and/or consisting of multiple types. A useful model should be versatile to address a variety of graph types, including bipartite graphs;
\item Incorporate side information: nodes (and maybe edges) are sometimes described by a set of features, called \emph{side information}, that could encode information complementary to topological features of the input graph. Such explicit data on nodes and edges are not always readily available and are considered \emph{optional}. A useful model should be able to incorporate optional side information about nodes and/or edges, whenever available, to potentially improve predictive performance;
\item Efficiency and scalability: real-world graph datasets contain large numbers of nodes and/or edges. It is essential for a model to be memory and computationally efficient to achieve practical utility on real-world applications.
\end{itemize}

\begin{figure*}
\centering
\includegraphics[width=\textwidth]{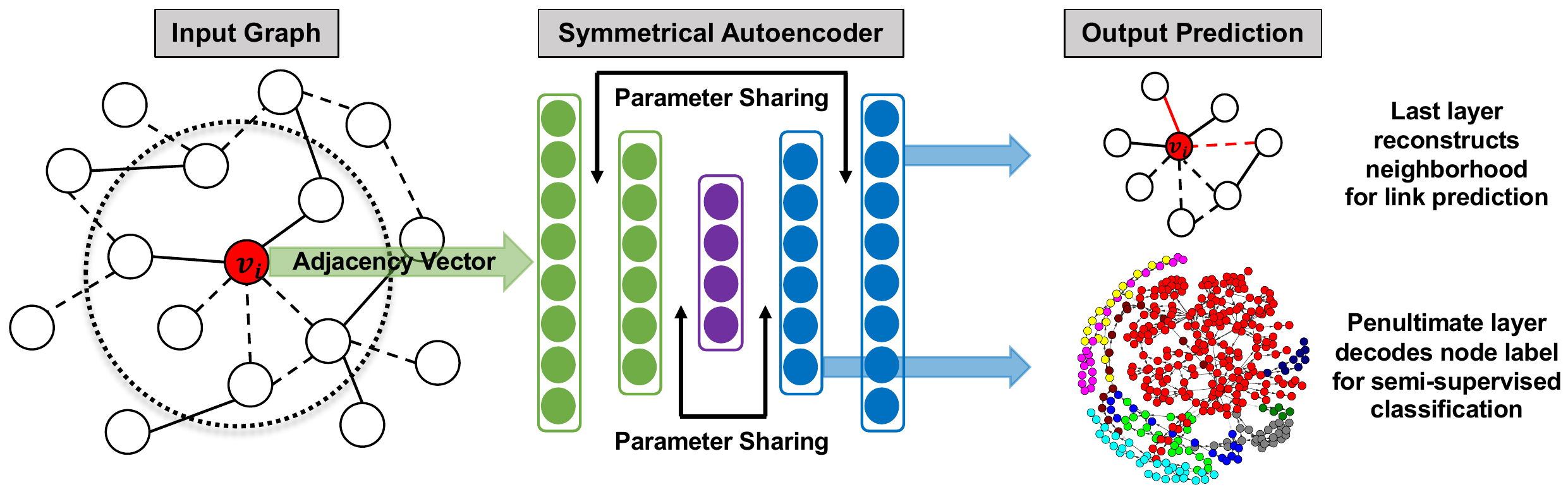}
\centering
\caption{Schematic depiction of the Local Neighborhood Graph Autoencoder (LoNGAE) architecture. \emph{Left}: A partially observed input graph with known positive links (solid lines) and known negative links (dashed lines) between two nodes; pairs of nodes not yet connected have unknown status links. \emph{Middle}: A symmetrical, densely connected autoencoder with parameter sharing (tied weights) is trained end-to-end to learn node embeddings from the adjacency vector for graph representation. \emph{Right}: Exemplar multi-task output for link prediction and node classification.}
\label{fig1}
\end{figure*}

Our contribution in this work is a simple, yet versatile autoencoder architecture that addresses all of the above technical challenges. We demonstrate that our autoencoder models: 1) can handle extreme class imbalance common in link prediction problems; 2) can learn expressive latent features for nodes from topological structures of sparse, bipartite graphs that may have directed and/or weighted edges; 3) is flexible to incorporate explicit side features about nodes as an optional component to improve predictive performance; and 4) utilize extensive parameter sharing to reduce memory footprint and computational complexity, while leveraging available GPU-based implementations for increased scalability. Further, the autoencoder architecture has the novelty of being efficiently trained end-to-end for the joint, multi-task learning (MTL) of both link prediction and node classification tasks. To the best of our knowledge, this is the first architecture capable of performing simultaneous link prediction and node classification in a single learning stage, whereas previous related methods require multiple training stages that are difficult to optimize. Lastly, we conduct a comprehensive evaluation of the proposed autoencoder architecture on nine challenging benchmark graph-structured datasets comprising a wide range of LPNC applications. Numerical experiments validate the efficacy of our models by showing significant improvement on multiple evaluation measures over related methods designed for link prediction and/or node classification.

\section{Autoencoder Architecture for Link Prediction and Node Classification}

We now characterize our proposed autoencoder architecture, schematically depicted in Figure~\ref{fig1}, for LPNC and formalize the notation used in this paper. The input to the autoencoder is a graph $\mathcal{G} = (\mathcal{V}, \mathcal{E})$ of $N = |\mathcal{V}|$ nodes. Graph $\mathcal{G}$ is represented by its adjacency matrix $\mathbf{A} \in \mathbb{R}^{N \times N}$. For a partially observed graph, $\mathbf{A} \in \{1,0,\textsc{unk}\}^{N \times N}$, where $1$ denotes a known present positive edge, $0$ denotes a known absent negative edge, and \textsc{unk} denotes an unknown status (missing or unobserved) edge. In general, the input to the autoencoder can be directed or undirected, weighted or unweighted, and/or bipartite graphs. However, for the remainder of this paper and throughout the numerical experiments, we assume undirected and symmetric graphs with binary edges to maintain parity with previous related work.

Optionally, we are given a matrix of available explicit node features, i.e. side information $\mathbf{X} \in \mathbb{R}^{N \times F}$. The aim of the autoencoder model $h(\mathbf{A,X})$ is to learn a set of low-dimensional latent variables for the nodes $\mathbf{Z} \in \mathbb{R}^{N \times D}$ that can produce an approximate reconstruction output $\mathbf{\hat{A}}$ such that the error between $\mathbf{A}$ and $\mathbf{\hat{A}}$ is minimized, thereby preserving the global graph structure. In this paper, we use capital variables (e.g., $\mathbf{A}$) to denote matrices and lower-case variables (e.g., $\mathbf{a}$) to denote row vectors. For example, we use $\mathbf{a}_i$ to mean the $i$th row of the matrix $\mathbf{A}$.

\subsection{Link Prediction}

Research on link prediction attempts to answer the principal question: given two entities, should there be a connection between them? We focus on the structural link prediction problem, where the task is to compute the likelihood that an unobserved or missing edge exists between two nodes in a partially observed graph. For a comprehensive survey on link prediction, to include \emph{structural} and \emph{temporal} link prediction using unsupervised and supervised models, see \cite{Wang:2014}. \\

\noindent \textbf{Link Prediction from Graph Topology} ~ Let $\mathbf{a}_i \in \mathbb{R}^N$ be an \emph{adjacency vector} of $\mathbf{A}$ that contains the local neighborhood of the $i$th node. Our proposed autoencoder architecture comprises a set of non-linear transformations on $\mathbf{a}_i$ summarized in two component parts: encoder $g(\mathbf{a}_i)\colon \mathbb{R}^N \to \mathbb{R}^D$, and decoder $f(\mathbf{z}_i)\colon \mathbb{R}^D \to \mathbb{R}^N$. We stack two layers of the encoder part to derive $D$-dimensional latent feature representation of the $i$th node $\mathbf{z}_i \in \mathbb{R}^D$, and then stack two layers of the decoder part to obtain an approximate reconstruction output $\mathbf{\hat{a}}_i \in \mathbb{R}^N$, resulting in a four-layer autoencoder architecture. Note that $\mathbf{a}_i$ is highly sparse, with up to 90 percent of the edges missing at random in some of our experiments, and the dense reconstructed output $\mathbf{\hat{a}}_i$ contains the predictions for the missing edges. The hidden representations for the encoder and decoder parts are computed as follows:
\begin{align*}
\mathbf{z}_i &= g\left(\mathbf{a}_i\right) = \text{ReLU}\left(\mathbf{W} \boldsymbol{\cdot} \text{ReLU}\left(\mathbf{V}\mathbf{a}_i + \mathbf{b}^{(1)}\right) + \mathbf{b}^{(2)}\right) \\ &\hspace{150pt} \text{Encoder Part}, \\
\mathbf{\hat{a}}_i &= f\left(\mathbf{z}_i\right) = \mathbf{V}^\text{T} \boldsymbol{\cdot} \text{ReLU}\left(\mathbf{W}^\text{T}\mathbf{z}_i + \mathbf{b}^{(3)}\right) + \mathbf{b}^{(4)} \\ &\hspace{150pt} \text{Decoder Part}, \\
\mathbf{\hat{a}}_i &= h\left(\mathbf{a}_i\right) = f\left(g\left(\mathbf{a}_i\right)\right) \\ &\hspace{150pt} \text{Autoencoder}.
\end{align*}

The choice of non-linear, element-wise activation function is the rectified linear unit $\text{ReLU}(\mathbf{x}) = \text{max}(0, \mathbf{x})$ \cite{Nair:2010}. The last decoder layer computes a linear transformation to score the missing links as part of the reconstruction. We constrain the autoencoder to be symmetrical with shared parameters for $\{\mathbf{W},\mathbf{V}\}$ between the encoder and decoder parts, resulting in almost $2\times$ fewer parameters than an unconstrained architecture. Parameter sharing is a powerful form of regularization that helps improve learning and generalization, and is also the main motivation for MTL, first explored in \cite{Caruana:1993} and most recently in \cite{Yang:2017}. Notice the bias units $\mathbf{b}$ do not share parameters, and $\big\{\mathbf{W}^\text{T}$, $\mathbf{V}^\text{T}\big\}$ are transposed copies of $\{\mathbf{W}$, $\mathbf{V}\}$. For brevity of notation, we summarize the parameters to be learned in $\theta = \big\{\mathbf{W}, \mathbf{V}, \mathbf{b}^{(k)}\big\}, k=1,...,4$. Since our autoencoder learns node embeddings from local neighborhood structures of the graph, we refer to it as LoNGAE for Local Neighborhood Graph Autoencoder.\\

\noindent \textbf{Link Prediction with Node Features} ~ Optionally, if a matrix of explicit node features $\mathbf{X} \in \mathbb{R}^{N \times F}$ is available, then we concatenate $(\mathbf{A},\mathbf{X})$ to obtain an \emph{augmented} adjacency matrix $\mathbf{\bar{A}} \in \mathbb{R}^{N \times (N + F)}$ and perform the above encoder-decoder transformations on $\mathbf{\bar{a}}_i$ for link prediction. We refer to this variant as $\alpha$LoNGAE. Notice the augmented adjacency matrix is no longer square and symmetric. The intuition behind the concatenation of node features is to enable a shared representation of both graph and node features throughout the autoencoding transformations by way of the tied parameters $\{\mathbf{W},\mathbf{V}\}$. This idea draws inspiration from recent work by Vukoti{\'c} et al. \cite{Vukotic:2016}, where they successfully applied symmetrical autoencoders with parameter sharing for multi-modal and cross-modal representation learning of textual and visual features.

The training complexity of $\alpha$LoNGAE is $\mathcal{O}((N+F)DI)$, where $N$ is the number of nodes, $F$ is the dimensionality of node features, $D$ is the size of the hidden layer, and $I$ is the number of iterations. In practice, $F$, $D \ll N$, and $I$ are independent of $N$. Thus, the overall complexity of the autoencoder is $\mathcal{O}(N)$, linear in the number of nodes.\\

\noindent \textbf{Inference and Learning} ~ During the forward pass, or inference, the model takes as input an adjacency vector $\mathbf{a}_i$ and computes its reconstructed output $\mathbf{\hat{a}}_i = h(\mathbf{a}_i)$ for link prediction. The parameters $\theta$ are learned via backpropagation. During the backward pass, we estimate $\theta$ by minimizing the Masked Balanced Cross-Entropy (MBCE) loss, which only allows for the contributions of those parameters associated with observed edges, as in \cite{Sedhain:2015}. Moreover, $\mathbf{a}_i$ can exhibit extreme class imbalance between known present and absent links, as is common in link prediction problems. We handle class imbalance by defining a weighting factor $\zeta \in [0,1]$ to be used as a multiplier for the positive class in the cross-entropy loss formulation. This approach is referred to as balanced cross-entropy. Other approaches to class imbalance include optimizing for a ranking loss \cite{Menon:2011} and the recent work on focal loss by Lin et al. \cite{Lin:2017}. For a single example $\mathbf{a}_i$ and its reconstructed output $\mathbf{\hat{a}}_i$, we compute the MBCE loss as follows:
\begin{align*}
\mathcal{L}_{\textsc{bce}} = -\mathbf{a}_i \log\left(\sigma\left(\mathbf{\hat{a}}_i\right)\right) \cdot \zeta - (1 - \mathbf{a}_i) &\log\left(1 - \sigma\left(\mathbf{\hat{a}}_i\right)\right), \\
\mathcal{L}_{\textsc{mbce}} = \frac{\mathbf{m}_i \odot \mathcal{L}_{\textsc{bce}}} {\sum \mathbf{m}_i}.
\end{align*}
Here, $\mathcal{L}_{\textsc{bce}}$ is the balanced cross-entropy loss with weighting factor $\zeta = 1 - \frac{\text{\# present links}}{\text{\# absent links}}$, $\sigma(\cdot)$ is the sigmoid function, $\odot$ is the Hadamard (element-wise) product,  and $\mathbf{m}_i$ is the boolean function: $\mathbf{m}_i = 1$  if $\mathbf{a}_i \neq \textsc{unk}$, else $\mathbf{m}_i = 0$.

The same autoencoder architecture can be applied to a row vector $\mathbf{\bar{a}}_i \in \mathbb{R}^{N+F}$ in the augmented adjacency matrix $\mathbf{\bar{A}}$. However, at the final decoder layer, we slice the reconstruction $h(\mathbf{\bar{a}}_i)$ into two outputs: $\mathbf{\hat{a}}_i \in \mathbb{R}^N$ corresponding to the reconstructed example in the original adjacency matrix, and $\mathbf{\hat{x}}_i \in \mathbb{R}^F$ corresponding to the reconstructed example in the matrix of node features. During learning, we optimize $\theta$ on the concatenation of graph topology and side node features $(\mathbf{a}_i, \mathbf{x}_i)$, but compute the losses for the reconstructed outputs $(\mathbf{\hat{a}}_i, \mathbf{\hat{x}}_i)$ separately with different loss functions. The motivation behind this design is to maintain flexibility to handle different input formats; the input $\mathbf{a}_i$ is usually binary, but the input $\mathbf{x}_i$ can be binary, real-valued, or both. In this work, we enforce both inputs $(\mathbf{a}_i, \mathbf{x}_i)$ to be in the range $[0, 1]$ for simplicity and improved performance, and compute the augmented $\alpha$MBCE loss as follows:
\begin{equation*}
\mathcal{L}_{\alpha \textsc{mbce}} = \mathcal{L}_{\textsc{mbce}(\mathbf{a}_i, \mathbf{\hat{a}}_i)} + \mathcal{L}_{\textsc{ce}(\mathbf{x}_i, \mathbf{\hat{x}}_i)},
\end{equation*}
where $\mathcal{L}_{\textsc{ce}(\mathbf{x}_i, \mathbf{\hat{x}}_i)} = -\mathbf{x}_i \log\left(\sigma\left(\mathbf{\hat{x}}_i\right)\right) - (1 - \mathbf{x}_i) \log\left(1 - \sigma\left(\mathbf{\hat{x}}_i\right)\right)$ is the standard cross-entropy loss with sigmoid function $\sigma(\cdot)$. At inference time, we use the reconstructed output $\mathbf{\hat{a}}_i$ for link prediction and disregard the output $\mathbf{\hat{x}}_i$.

\subsection{Semi-Supervised Node Classification}

The $\alpha$LoNGAE model can also be used to perform efficient information propagation on graphs for the task of semi-supervised node classification. Node classification is the task of predicting the labels or types of entities in a graph, such as the types of molecules in a metabolic network or document categories in a citation network.

For a given augmented adjacency vector $\mathbf{\bar{a}}_i$, the autoencoder learns the corresponding node embeddings $\mathbf{z}_i$ to obtain an optimal reconstruction. Intuitively, $\mathbf{z}_i$ encodes a vector of latent features derived from the concatenation of both graph and node features, and can be used to predict the label of the $i$th node. For multi-class classification, we can decode $\mathbf{z}_i$ using the softmax activation function to learn a probability distribution over node labels. More precisely, we predict node labels via the following transformation: $\mathbf{\hat{y}}_i = \text{softmax}(\mathbf{\tilde{z}}_i) = \frac{1}{\mathcal{Z}} \exp(\mathbf{\tilde{z}}_i)$, where $\mathcal{Z} = \sum \exp(\mathbf{\tilde{z}}_i)$ and $\mathbf{\tilde{z}}_i = \mathbf{U} \boldsymbol{\cdot} \text{ReLU}\left(\mathbf{W}^\text{T} \mathbf{z}_i + \mathbf{b}^{(3)}\right) + \mathbf{b}^{(5)}$.

In many applications, only a small fraction of the nodes are labeled. For semi-supervised learning, it is advantageous to utilize unlabeled examples in conjunction with labeled instances to better capture the underlying data patterns for improved learning and generalization. We achieve this by jointly training the autoencoder with a masked softmax classifier to collectively learn node labels from minimizing their combined losses:
\begin{equation*}
\mathcal{L}_{\textsc{multi-task}} = - \textsc{mask}_i \sum_{c \in C}\mathbf{y}_{ic} \log (\mathbf{\hat{y}}_{ic}) + \mathcal{L}_{\textsc{mbce}},
\end{equation*}
where $C$ is the set of node labels, $\mathbf{y}_{ic} = 1$ if node $i$ belongs to class $c$, $\mathbf{\hat{y}}_{ic}$ is the softmax probability that node $i$ belongs to class $c$, $\mathcal{L}_{\textsc{mbce}}$ is the loss defined for the autoencoder, and the boolean function $\textsc{mask}_i = 1$ if node $i$ has a label, otherwise $\textsc{mask}_i = 0$. Notice in this configuration, we can perform multi-task learning for both link prediction and semi-supervised node classification, \emph{simultaneously}.

\section{Related Work}

The field of graph representation learning is seeing a resurgence of research interest in recent years, driven in part by the latest advances in deep learning. The aim is to learn a mapping that encodes the input graph into low-dimensional feature embeddings while preserving its original global structure. Hamilton et al. \cite{Hamilton:2017} succinctly articulate the diverse set of previously proposed approaches for graph representation learning, or graph embedding, as belonging within a unified encoder-decoder framework. In this section, we summarize three classes of encoder-decoder models most related to our work: matrix factorization (MF), autoencoders, and graph convolutional networks (GCNs).

MF has its roots in dimensionality reduction and gained popularity with extensive applications in collaborative filtering (CF) and recommender systems \cite{Koren:2009}. MF models take an input matrix $\mathbf{M}$, learn a shared linear latent representation for rows ($\mathbf{r}_i$) and columns ($\mathbf{c}_j$) during an encoder step, and then use a bilinear (pairwise) decoder based on the inner product $\mathbf{r}_i\mathbf{c}_j$ to produce a reconstructed matrix $\mathbf{\hat{M}}$. CF is mathematically similar to link prediction, where the goal is essentially matrix completion. Menon and Elkan \cite{Menon:2011} proposed an MF model capable of incorporating side information about nodes and/or edges to demonstrate strong link prediction results on several challenging network datasets. Other recent approaches similar to MF that learn node embeddings via some encoder transformation and then use a bilinear decoder for the reconstruction include DeepWalk \cite{Perozzi:2014} and its variants LINE \cite{Tang:2015} and node2vec \cite{Grover:2016}. DeepWalk, LINE, and node2vec do not support external node/edge features.

Our work is inspired by recent successful applications of autoencoder architectures for collaborative filtering that outperform popular matrix factorization methods \cite{Sedhain:2015,Strub:2016,Kuchaiev:2017}, and is related to Structural Deep Network Embedding (SDNE) \cite{Wang:2016} for link prediction. Similar to SDNE, our models rely on the autoencoder to learn non-linear node embeddings from local graph neighborhoods. However, our models have several important distinctions: 1) we leverage extensive parameter sharing between the encoder and decoder parts to enhance representation learning; 2) our $\alpha$LoNGAE model can optionally concatenate side node features to the adjacency matrix for improved link prediction performance; and 3) the $\alpha$LoNGAE model can be trained end-to-end in a single stage for multi-task learning of link prediction and semi-supervised node classification. On the other hand, training SDNE requires multiple steps that are difficult to jointly optimize: i) pre-training via a deep belief network; and ii) utilizing a separate downstream classifier on top of node embeddings for LPNC.

Lastly, GCNs \cite{Kipf:2016} are a recent class of algorithms based on convolutional encoders for learning node embeddings. The GCN model is motivated by a localized first-order approximation of spectral convolutions for layer-wise information propagation on graphs. Similar to our $\alpha$LoNGAE model, the GCN model can learn hidden layer representations that encode both local graph structure and features of nodes. The choice of the decoder depends on the task. For link prediction, the bilinear inner product is used in the context of the variational graph autoencoder (VGAE) \cite{VGAE:2016}. For semi-supervised node classification, the softmax activation function is employed. The GCN model provides an end-to-end learning framework that scales linearly in the number of graph edges and has been shown to achieve strong LPNC results on a number of graph-structured datasets. However, the GCN model has a drawback of being memory intensive because it is trained on the full dataset using batch gradient descent for every training iteration. We show that our models outperform GCN-based models for LPNC while consuming a constant memory budget by way of mini-batch training.

\section{Experimental Design}
In this section, we expound our protocol for the empirical evaluation of our models' capability for learning and generalization on the tasks of link prediction and semi-supervised node classification. Secondarily, we also present results of the models' representation capacity on the task of network reconstruction.

\subsection{Datasets and Baselines}
We evaluate our proposed autoencoder models on nine graph-structured datasets, spanning multiple application domains, from which previous graph embedding methods have achieved strong results for LPNC. The datasets are summarized in Table~\ref{tab1} and include networks for \texttt{Protein} interactions, \texttt{Metabolic} pathways, military \texttt{Conflict} between countries, the U.S. \texttt{PowerGrid}, collaboration between users on the \texttt{BlogCatalog} social website, and publication citations from the \texttt{Cora}, \texttt{Citeseer}, \texttt{Pubmed}, \texttt{Arxiv-GRQC} databases. \texttt{\{Protein, Metabolic, Conflict, PowerGrid\}} are reported in \cite{Menon:2011}. \texttt{\{Cora, Citeseer, Pubmed\}} are from \cite{Sen:2008} and reported in \cite{Kipf:2016,VGAE:2016}. And \texttt{\{Arxiv-GRQC, BlogCatalog\}} are reported in \cite{Wang:2016}.

\begin{table}[ht]
\begin{center}
\caption[Caption for Table 1]{Summary statistics of datasets used in empirical evaluation. The notation $\vert\mathit{O}^+\vert$:$\vert\mathit{O}^-\vert$ denotes the ratio of observed present (positive) edges to absent (negative) edges and is a measure of class imbalance. Label rate is defined as the number of nodes labeled for training divided by the total number of nodes.}
\begin{adjustbox}{width=0.5\textwidth}
	\begin{tabular} {l  r  r  r  r  r  r  r  r}
	\hline
	\multicolumn{1}{l}{\multirow{2}{*}{\textbf{Dataset}}} &
	\multicolumn{1}{c}{\multirow{2}{*}{\textbf{Nodes}}} &
    \multicolumn{1}{c}{\multirow{1}{*}{\textbf{Average}}} &
	\multicolumn{1}{c}{\multirow{1}{*}{\textbf{$\vert\mathit{O}^+\vert$:$\vert\mathit{O}^-\vert$}}} &
	\multicolumn{1}{c}{\multirow{1}{*}{\textbf{Node}}} &
	\multicolumn{1}{c}{\multirow{1}{*}{\textbf{Node}}} &
	\multicolumn{1}{c}{\multirow{1}{*}{\textbf{Label}}} \\
    & {}
    & \multicolumn{1}{c}{\multirow{1}{*}{\textbf{Degree}}}
    & \multicolumn{1}{c}{\multirow{1}{*}{\textbf{Ratio}}}
    & \multicolumn{1}{c}{\multirow{1}{*}{\textbf{Features}}}
    & \multicolumn{1}{c}{\multirow{1}{*}{\textbf{Classes}}}
    & \multicolumn{1}{c}{\multirow{1}{*}{\textbf{Rate}}} \\
    \hline \hline
    Pubmed
				& 19,717
				& 4.5
                & $1:4384$
				& 500
				& 3
				& 0.003 \\
    Citeseer
				& 3,327 
				& 2.8
                & $1:1198$
				& 3,703
				& 6
				& 0.036 \\
	Cora	
				& 2,708
				& 3.9
                & $1: \hspace{4.5pt} 694$
				& 1,433
				& 7
				& 0.052 \\
	Protein
					& 2,617
					& 9.1
                    & $1: \hspace{4.5pt} 300$
					& 76
					& --
					& -- \\
	Metabolic
					& 668
					& 8.3
                    & $1: \hspace{9pt} 80$
					& 325
					& --
					& -- \\
	Conflict
					& 130
					& 2.5
                    & $1: \hspace{9pt} 52$
					& 3
					& --
					& -- \\
	PowerGrid
				 & 4,941
				 & 2.7
                 & $1:1850$
				 & --
				 & --
				 & -- \\
    Arxiv-GRQC
    		    & 5,242
                & 5.5
                & $1: \hspace{4.5pt} 947$
                & --
                & --
                & -- \\
	BlogCatalog
    		    & 10,312
                & 64.8
                & $1: \hspace{4.5pt} 158$
                & --
                & --
                & -- \\
	\hline
	\end{tabular}
	\label{tab1}
\end{adjustbox}
\end{center}
\end{table}

\begin{table}[ht]
\begin{center}
\caption[Caption for Table 2]{Summary of baselines used in empirical evaluation. Acronyms: AUC -- Area Under ROC Curve; AP -- Average Precision.}
\begin{adjustbox}{width=0.5\textwidth}
	\begin{tabular} {l  l  l}
	\hline
	\multicolumn{1}{l}{\multirow{1}{*}{\textbf{Baseline}}} &
	\multicolumn{1}{l}{\multirow{1}{*}{\textbf{Task}}} &
	\multicolumn{1}{l}{\multirow{1}{*}{\textbf{~~Metric}}} \\ \hline \hline
    SDNE \cite{Wang:2016} ~
							& Reconstruction
							& ~~Precision@$k$ \\
    MF \cite{Menon:2011} ~
					& Link Prediction
					 & ~~AUC \\
    VGAE \cite{VGAE:2016} ~
				& Link Prediction
				 & ~~AUC, AP \\
    GCN \cite{Kipf:2016} ~
						& Node Classification
				 		& ~~Accuracy \\
	\hline
	\end{tabular}
	\label{tab2}
\end{adjustbox}
\end{center}
\end{table}

We empirically compare our autoencoder models against four strong baselines summarized in Table~\ref{tab2}, which were designed specifically for link prediction and/or node classification. We begin our empirical evaluation with the SDNE \cite{Wang:2016} baseline, where we compare the representation capacity of our models on the network reconstruction task using the \texttt{Arxiv-GRQC} and \texttt{BlogCatalog} datasets. For the MF baseline, we closely follow the experimental protocol in \cite{Menon:2011}, where we randomly sample 10 percent of the observed links for training and evaluate link prediction performance on the other disjoint 90 percent for the \texttt{\{Protein, Metabolic, Conflict\}} datasets. For \texttt{PowerGrid}, we use 90 percent of observed links for training and evaluate on the remaining 10 percent. And for the VGAE and GCN baselines, we use the same train/validation/test segments described in \cite{VGAE:2016} and \cite{Kipf:2016} for link prediction and node classification, respectively, on the \texttt{\{Cora, Citeseer, Pubmed\}} citation networks.

\subsection{Implementation Details}
We implement the autoencoder architecture using Keras \cite{Keras} on top of the GPU-enabled TensorFlow \cite{TF} backend, along with several additional details. The diagonal elements of the adjacency matrix are set to $1$ with the interpretation that every node is connected to itself. We impute missing or \textsc{unk} elements in the adjacency matrix with $0$. Note that imputed edges are not observed elements in the adjacency matrix and hence do not contribute to the masked loss computations during training. We are free to impute any values for the missing edges, but through cross-validation we found that the uniform value of $0$ produces the best results.

Hyper-parameter tuning is performed via cross-validation or on the available validation set. Key hyper-parameters include mini-batch size, dimensionality of the hidden layers, and the percentage of dropout regularization. In general, we strive to keep a similar set of hyper-parameters across datasets to highlight the consistency of our models. In all experiments, the dimensionality of the hidden layers in the autoencoder architecture is fixed at $N$-256-128-256-$N$. For reconstruction and link prediction, we train for 50 epochs using mini-batch size of 8 samples. For node classification, we train for 100 epochs using mini-batch size of 64 samples. We utilize early stopping as a form of regularization in time when the model shows signs of overfitting on the validation set.

We apply mean-variance normalization (MVN) after each ReLU activation layer to help improve link prediction performance, where it compensates for noise between train and test instances by normalizing the activations to have zero mean and unit variance. MVN enables efficient learning and has been shown effective in cardiac semantic segmentation \cite{Tran:2016} and speech recognition \cite{Joshi:2016}.

During training, we apply dropout regularization \cite{Srivastava:2014} throughout the architecture to mitigate overfitting, depending on the sparsity of the input graph. For link prediction, dropout is also applied at the input layer to produce an effect similar to using a \emph{denoising} autoencoder. This denoising technique was previously employed for link prediction in \cite{Chen:2014}. We initialize weights according to the Xavier scheme described in \cite{Xavier:2010}. We do not apply weight decay regularization.

We employ the Adam algorithm \cite{Kingma:2015} for gradient descent optimization with a fixed learning rate of $0.001$. As part of our experimental design, we also performed experiments without parameter sharing between the encoder and decoder parts of the architecture and found severely degraded predictive performance. This observation is consistent with prior findings that parameter sharing helps improve generalization by providing additional regularization to mitigate the adverse effects of overfitting and enhance representation learning \cite{Vukotic:2016,Yang:2017}.
\subsection{Results and Analysis}
\begin{figure*}
\centering
\includegraphics[width=0.9\textwidth]{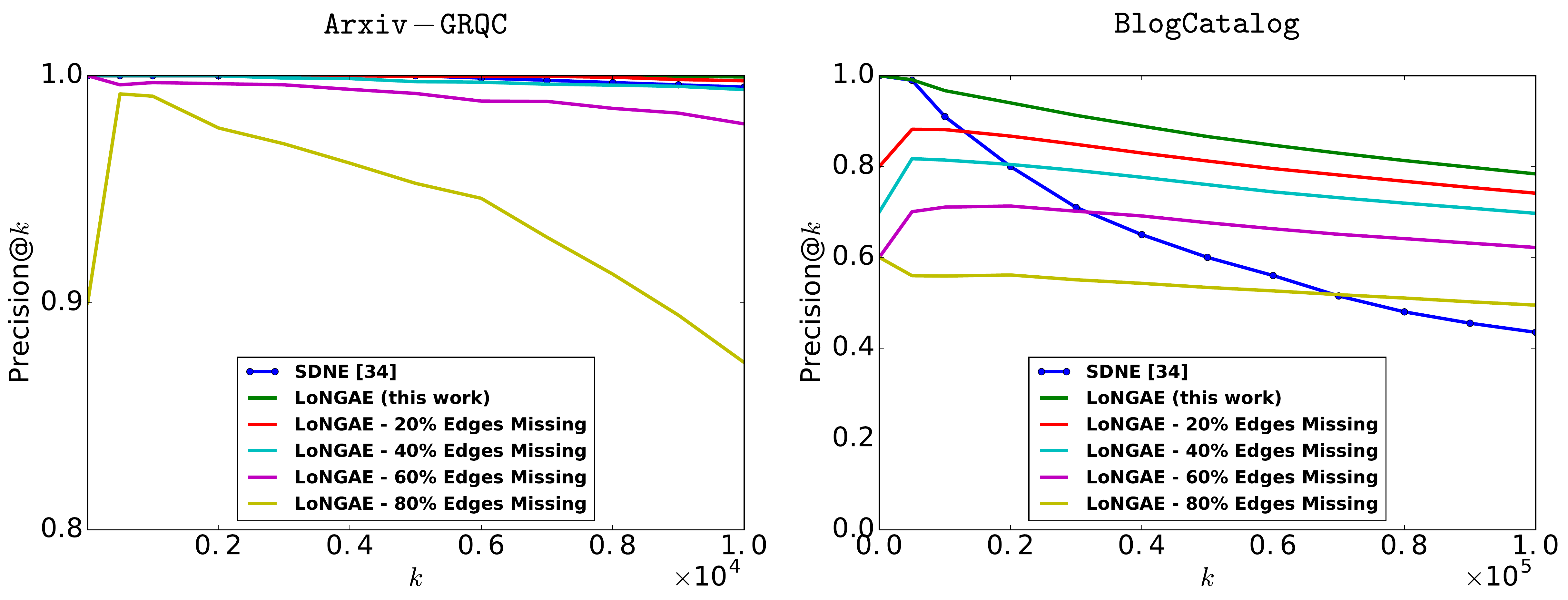}
\centering
\caption{Comparison of precision@$k$ performance between our LoNGAE model and the related autoencoder-based SDNE model for the reconstruction task on the \texttt{Arxiv-GRQC} and \texttt{BlogCatalog} network datasets. The parameter $k$ indicates the total number of retrieved edges.}
\label{fig2}
\end{figure*}

\noindent \textbf{Reconstruction} ~~ Results of the reconstruction task for the \texttt{Arxiv-GRQC} and \texttt{BlogCatalog} network datasets are illustrated in Figure~\ref{fig2}. In this experiment, we compare the results obtained by our LoNGAE model to those obtained by the related autoencoder-based SDNE model \cite{Wang:2016}. The evaluation metric is precision@$k$, which is a rank-based measure used in information retrieval and is defined as the proportion of retrieved edges/links in the top-$k$ set that are relevant. We use precision@$k$ to evaluate the model's ability to retrieve edges known to be present (positive edges) as part of the reconstruction.

\begin{table*}[ht]
\begin{center}
\caption[Caption for Table 3]{Comparison of AUC performance between our autoencoder models and the best previous matrix factorization model for link prediction. Number format: mean value (standard deviation). \footnotemark[1]{These results incorporated additional \emph{edge features} for link prediction, which we leave for future work.}}
\begin{adjustbox}{width=0.8\textwidth}
	\begin{tabular} {l  c  r  r  r  r}
	\hline
	\multicolumn{1}{l}{\multirow{1}{*}{\textbf{Method}}} &
	\multicolumn{1}{c}{\multirow{1}{*}{\textbf{Node Features}}} &
	\multicolumn{1}{c}{\multirow{1}{*}{\textbf{Protein}}} &
	\multicolumn{1}{c}{\multirow{1}{*}{\textbf{Metabolic}}} &
	\multicolumn{1}{c}{\multirow{1}{*}{\textbf{Conflict}}} &
	\multicolumn{1}{c}{\multirow{1}{*}{\textbf{PowerGrid}}} \\ \hline \hline
	LoNGAE (this work)
							& No
							& \textbf{0.798 (0.004)}
							& \textbf{0.703 (0.009)}
							& \textbf{0.698 (0.025)}
							& \textbf{0.781 (0.007)} \\
	Matrix Factorization \cite{Menon:2011}
					& No
					 & 0.795 (0.005)
					 & 0.696 (0.001)
					 & 0.692 (0.040)
					 & 0.754 (0.014) \\
	\hline
	$\alpha$LoNGAE (this work)
						& Yes
				 		& \textbf{0.861 (0.003)}
						& 0.750 (0.011)
						& 0.699 (0.021)
						& -- \\
	Matrix Factorization \cite{Menon:2011}
				& Yes
				 & 0.813 (0.002)
				 & \footnotemark[1]}\textbf{0.763 (0.006)
				 & \footnotemark[1]}\textbf{0.890 (0.017)
				 & -- \\
	
	\hline
	\end{tabular}
	\label{tab3}
\end{adjustbox}
\end{center}
\end{table*}

In comparison to SDNE, we show that our LoNGAE model achieves better precision@$k$ performance for all $k$ values, up to $k=10,000$ for \texttt{Arxiv-GRQC} and $k=100,000$ for \texttt{BlogCatalog}, when trained on the complete datasets. We also systematically test the capacity of the LoNGAE model to reconstruct the original networks when up to 80 percent of the edges are randomly removed, akin to the link prediction task. We show that the LoNGAE model only gets worse precision@$k$ performance than SDNE on the \texttt{Arxiv-GRQC} dataset when more than 40 percent of the edges are missing at random. On the \texttt{BlogCatalog} dataset, the LoNGAE model achieves better precision@$k$ performance than SDNE for large $k$ values even when 80 percent of the edges are missing at random. This experiment demonstrates the superior representation capacity of our LoNGAE model compared to SDNE. \\

\noindent \textbf{Link Prediction} ~ Table~\ref{tab3} shows the comparison between our autoencoder models and the matrix factorization (MF) model proposed in \cite{Menon:2011} for link prediction with and without node features. Recall that our goal is to recover the statuses of the missing or unknown links in the input graph. As part of the experimental design, we pretend that a randomly selected set of elements in the adjacency matrix are missing and collect their indices to be used as a validation set. Our task is to train the autoencoder to produce a set of predictions, a list of ones and zeros, on those missing indices and see how well the model performs when compared to the ground-truth. The evaluation metric is the area under the ROC curve (AUC). Results are reported as mean AUC and standard deviation over 10-fold cross-validation. The datasets under consideration for link prediction exhibit varying degrees of class imbalance.

For featureless link prediction, our LoNGAE model marginally outperforms MF on \texttt{\{Protein, Metabolic, Conflict\}} and is significantly better than MF on \texttt{PowerGrid}. Consistent with MF results, we observe that incorporating external node features provides a boost in link prediction accuracy, especially for the \texttt{Protein} dataset where we achieve a 6 percent increase in performance. \texttt{Metabolic} and \texttt{Conflict} also come with external \emph{edge features}, which were exploited by the MF model for further performance gains. We leave the task of combining edge features for future work. Each node in \texttt{Conflict} only has three features, which are unable to significantly boost link prediction accuracy. \texttt{PowerGrid} does not have node features so there are no results for the respective rows.

Table~\ref{tab4} summarizes the performances between our autoencoder models and related graph embedding methods for link prediction with and without node features. Following the protocol described in \cite{VGAE:2016}, we report AUC and average precision (AP) scores for each model on the held-out test set containing 10 percent of randomly sampled positive links and the same number of negative links. We show mean AUC and AP with standard error over 10 runs with random weight initializations on fixed data splits. Results for the baseline methods are taken from Kipf and Welling \cite{VGAE:2016}, where we pick the best performing models for comparison. Similar to the MF model, the graph embedding methods that can combine side node features \emph{always} produce a boost in link prediction accuracy. In this comparison, we significantly outperform the best graph embedding methods by as much as 10 percent, with and without node features. Our $\alpha$LoNGAE model achieves competitive link prediction performance when compared against the best model presented in \cite{VGAE:2016} on the \texttt{Pubmed} dataset.

\begin{table*}[ht]
\begin{center}
\caption[Caption of Table 4]{Comparison of AUC and AP performances between our autoencoder models and related graph embedding methods for link prediction. Number format: mean value (standard deviation). \footnotemark[4]Denotes the best performing model presented in \cite{VGAE:2016}.}
\begin{adjustbox}{width=0.9\textwidth}
	\begin{tabular} {l  c  c  c  c  c  c  c}
	\hline
	\multicolumn{1}{l}{\multirow{2}{*}{\textbf{Method}}} &
	\multicolumn{1}{c}{\multirow{1}{*}{\textbf{Node}}} & 
	\multicolumn{2}{c}{\multirow{1}{*}{\textbf{Cora}}} &
	\multicolumn{2}{c}{\multirow{1}{*}{\textbf{Citeseer}}} &
	\multicolumn{2}{c}{\multirow{1}{*}{\textbf{Pubmed}}} \\
	& \textbf{Features} & \textbf{AUC} & \textbf{AP} & \textbf{AUC} & \textbf{AP} & \textbf{AUC} & \textbf{AP} \\ \hline \hline
	LoNGAE (this work)
							& No
							& \textbf{0.896 (0.003)}
							& \textbf{0.915 (0.001)}
							& \textbf{0.860 (0.003)}
							& \textbf{0.892 (0.003)}
							& \textbf{0.926 (0.001)}
							& \textbf{0.930 (0.002)} \\
	Spectral Clustering \cite{Tang:2011}
					& No
					 & 0.846 (0.01)
					 & 0.885 (0.00)
					 & 0.805 (0.01)
					 & 0.850 (0.01)
					 & 0.842 (0.01)
					 & 0.878 (0.01) \\
	DeepWalk \cite{Perozzi:2014}
						& No
					        & 0.831 (0.01)
					        & 0.850 (0.00)
					        & 0.805 (0.02)
					        & 0.836 (0.01)
					        & 0.844 (0.00)
					        & 0.841 (0.00) \\
	VGAE\footnotemark[4] \cite{VGAE:2016}
					& No
				        & 0.843 (0.02)
				        & 0.881 (0.01)
				        & 0.789 (0.03)
				        & 0.841 (0.02)
				        & 0.827 (0.01)
				        & 0.875 (0.01) \\
	\hline
	$\alpha$LoNGAE (this work)
						& Yes
				 		& \textbf{0.943 (0.003)}
						& \textbf{0.952 (0.002)}
						& \textbf{0.956 (0.003)}
						& \textbf{0.964 (0.002)}
						& 0.960 (0.003)
						& 0.963 (0.002) \\
	VGAE\footnotemark[4] \cite{VGAE:2016}
				& Yes
				 & 0.914 (0.01)
				 & 0.926 (0.01)
				 & 0.908 (0.02)
				 & 0.920 (0.02)
				 & \textbf{0.964 (0.00)}
				 & \textbf{0.965 (0.00)} \\
	
	\hline
	\end{tabular}
	\label{tab4}
\end{adjustbox}
\end{center}
\end{table*}

\noindent \textbf{Node Classification} ~ Results of semi-supervised node classification for the \texttt{\{Cora, Citeseer, Pubmed\}} datasets are summarized in Table~\ref{tab5}. In this context of citation networks, node classification is equivalent to the task of document classification. We closely follow the experimental setup of Kipf and Welling \cite{Kipf:2016}, where we use their provided train/validation/test splits for evaluation. Accuracy performance is measured on the held-out test set of 1,000 examples. We tune hyper-parameters on the validation set of 500 examples. The train set only contains 20 examples per class. All methods use the complete adjacency matrix, and available node features, to learn latent embeddings for node classification. For comparison, we train and test our $\alpha$LoNGAE model on the same data splits over 10 runs with random weight initializations and report mean accuracy. Kipf and Welling \cite{Kipf:2016} report their mean GCN and ICA results on the same data splits over 100 runs with random weight initializations. The other baseline methods are taken from Yang et al. \cite{Yang:2016}. In this comparison, our $\alpha$LoNGAE model achieves competitive performance when compared against the GCN model on the \texttt{Cora} dataset, but outperforms GCN and all other baseline methods on the \texttt{Citeseer} and \texttt{Pubmed} datasets.

\begin{table}[ht]
\begin{center}
\caption[Caption for Table 5]{Comparison of accuracy performance between our $\alpha$LoNGAE model and related graph embedding methods for semi-supervised node classification.}
\begin{adjustbox}{width=0.5\textwidth}
	\begin{tabular} {l  r  r  r}
	\hline
	\multicolumn{1}{l}{\multirow{1}{*}{\textbf{Method}} } &
	\multicolumn{1}{r}{\multirow{1}{*}{\textbf{Cora}}} &
	\multicolumn{1}{r}{\multirow{1}{*}{\textbf{~Citeseer}}} &
	\multicolumn{1}{r}{\multirow{1}{*}{\textbf{~Pubmed}}} \\ \hline \hline
	$\alpha$LoNGAE (this work) ~
							& 0.783
							& \textbf{0.716}
							& \textbf{0.794} \\
	GCN \cite{Kipf:2016} 
					 & \textbf{0.815}
					 & 0.703
					 & 0.790 \\
	Planetoid \cite{Yang:2016}
				 		& 0.757
						& 0.647
						& 0.772 \\
	ICA \cite{Lu:2003}
				 & 0.751
				 & 0.691
				 & 0.739 \\
	DeepWalk \cite{Perozzi:2014}
				& 0.672
				& 0.432
				& 0.653 \\
	\hline
	\end{tabular}
	\label{tab5}
\end{adjustbox}
\end{center}
\end{table}

\noindent \textbf{Multi-task Learning} ~~ Lastly, we report LPNC results obtained by our $\alpha$LoNGAE model in the MTL setting over 10 runs with random weight initializations. In the MTL scenario, the $\alpha$LoNGAE model takes as input an incomplete graph with 10 percent of the positive edges, and the same number of negative edges, missing at random and all available node features to simultaneously produce predictions for the missing edges and labels for the nodes. Table~\ref{tab6} shows the efficacy of the $\alpha$LoNGAE model for MTL when compared against the best performing task-specific link prediction and node classification models, which require the complete adjacency matrix as input. For link prediction, multi-task $\alpha$LoNGAE achieves competitive performance against task-specific $\alpha$LoNGAE, and significantly outperforms the best VGAE model from Kipf and Welling \cite{VGAE:2016} on \texttt{Cora} and \texttt{Citeseer} datasets. For node classification, multi-task $\alpha$LoNGAE is the best performing model across the board, only trailing behind the GCN model on the \texttt{Cora} dataset.

\begin{table}[ht]
\begin{center}
\caption[Caption for Table 6]{Comparison of link prediction and node classification performances obtained by the $\alpha$LoNGAE model in the multi-task learning setting. Link prediction performance is reported as the combined average of AUC and AP scores. Accuracy is used for node classification performance.}
\begin{adjustbox}{width=0.5\textwidth}
	\begin{tabular} {l  r  r  r}
	\hline
	\multicolumn{1}{l}{\multirow{1}{*}{\textbf{Method}} } &
	\multicolumn{1}{r}{\multirow{1}{*}{\textbf{Cora}}} &
	\multicolumn{1}{r}{\multirow{1}{*}{\textbf{~Citeseer}}} &
	\multicolumn{1}{r}{\multirow{1}{*}{\textbf{~Pubmed}}} \\ \hline \hline
    \multicolumn{4}{c}{\multirow{1}{*}{\textbf{Link Prediction}}} \\
	Multi-task $\alpha$LoNGAE ~
						& 0.946
						& 0.949
						& 0.944 \\
	Task-specific $\alpha$LoNGAE ~
				 & \textbf{0.948}
				 & \textbf{0.960}
				 & 0.962 \\
	Task-specific VGAE \cite{VGAE:2016} ~
				 & 0.920
				 & 0.914
				 & \textbf{0.965} \\
	\hline \hline
    \multicolumn{4}{c}{\multirow{1}{*}{\textbf{Node Classification}}} \\
	Multi-task $\alpha$LoNGAE
							& 0.790
							& \textbf{0.718}
							&  \textbf{0.804} \\
	Task-specific $\alpha$LoNGAE
							& 0.783
							& 0.716
							& 0.794 \\
	Task-specific GCN \cite{Kipf:2016} 
					 & \textbf{0.815}
					 & 0.703
					 & 0.790 \\
	\hline
	\end{tabular}
	\label{tab6}
\end{adjustbox}
\end{center}
\end{table}

\section{Discussion}
In our experiments, we show that a simple autoencoder architecture with parameter sharing consistently outperforms previous related methods on a range of challenging graph-structured benchmarks for three separate tasks: reconstruction, link prediction, and semi-supervised node classification. For the reconstruction task, our LoNGAE model achieves superior precision@$k$ performance when compared to the related SDNE model. Although both models leverage a deep autoencoder architecture for graph representation learning, the SDNE model lacks several key implementations necessary for enhanced representation capacity, namely parameter sharing between the encoder-decoder parts and end-to-end training of deep architectures.

For link prediction, we observe that combining available node features \emph{always} produces a significant boost in predictive performance. This observation was previously reported in \cite{Menon:2011,VGAE:2016}, among others. Intuitively, we expect topological graph features provide complementary information not present in the node features, and the combination of both feature sets should improve predictive power. Although explicit node features may not always be readily available, a link prediction model capable of incorporating optional side information has broader applicability.

Our $\alpha$LoNGAE model also performs favorably well on the task of semi-supervised node classification. The model is capable of encoding non-linear node embeddings from both local graph structure and explicit node features, which can be decoded by a softmax activation function to yield accurate node labels. The efficacy of the proposed $\alpha$LoNGAE model is evident especially on the \texttt{Pubmed} dataset, where the label rate is only 0.003. This efficacy is attributed to parameter sharing being used in the autoencoder architecture, which provides regularization to help improve representation learning and generalization.

Our autoencoder architecture naturally supports multi-task learning, where a joint representation for both link prediction and node classification is enabled via parameter sharing. MTL aims to exploit commonalities and differences across multiple tasks to find a shared representation that can result in improved performance for each task-specific metric. In this work, we show that our multi-task $\alpha$LoNGAE model improves node classification accuracy by learning to predict missing edges at the same time. Our multi-task model has broad practical utility to address real-world applications where the input graphs may have both missing edges and node labels.

Finally, we address one major limitation associated with our autoencoder models having complexity scale linearly in the number of nodes. Hamilton et al. \cite{Hamilton:2017} express that the complexity in nodes may limit the utility of the models on massive graphs with hundreds of millions of nodes. In practice, we would implement our models to leverage data parallelism \cite{Shrivastava:2017} across commodity CPU and/or GPU resources for effective distributed learning on massive graphs. Data parallelism is possible because our models learn node embeddings from each row vector of the adjacency matrix independently. Nevertheless, the area of improvement in future work is to take advantage of the sparsity of edges in the graphs to scale our models linearly in the number of observed edges.

\section{Conclusion}
In this work, we presented a new autoencoder architecture for link prediction and semi-supervised node classification, and showed that the resulting models outperform related methods in accuracy performance on a range of real-world graph-structured datasets. The success of our models is primarily attributed to extensive parameter sharing between the encoder and decoder parts of the architecture, coupled with the capability to learn expressive non-linear latent node representations from both local graph neighborhoods and explicit node features. Further, our novel architecture is capable of simultaneous multi-task learning of both link prediction and node classification in one efficient end-to-end training stage. Our work provides a useful framework to make accurate and meaningful predictions on a diverse set of complex graph structures for a wide range of real-world applications.

\section*{Acknowledgment}
The author thanks Edward Raff and Jared Sylvester for insightful discussions, and gracious reviewers for constructive feedback on the paper.


%
%

\end{document}